\documentclass[letterpaper, 10 pt, conference]{ieeeconf}  

\IEEEoverridecommandlockouts                              

\usepackage[letterpaper,
            top=0.75in,
            bottom=1in,
            left=0.625in,
            right=0.625in,
            columnsep=0.25in]{geometry}

\usepackage[utf8]{inputenc} 
\usepackage[T1]{fontenc}    
\usepackage{hyperref}       
\hypersetup{pdftitle={Enhancing Computer Vision Model Generalization in Warehouse Facilities: 
A Case Study on Anomaly Detection in Vertical Material Handling Systems}, pdfauthor={Ruiliang Liu, Tina Dongxu Li, Joshua Migdal, Ken Meszaros, Trevor Dardik}}
\usepackage{url}            
\usepackage{booktabs}       
\usepackage{amsfonts}       
\usepackage{nicefrac}       
\usepackage{microtype}      
\usepackage{graphicx}

\title{Enhancing Computer Vision Model Generalization in Warehouse Facilities: 
A Case Study on Anomaly Detection in Vertical Material Handling Systems
}

\author{Ruiliang Liu, Tina Dongxu Li, Joshua Migdal, Ken Meszaros, Trevor Dardik \\
Amazon, USA\\
\{ruilianl, dxl, jmigdal, mesza, tdardik\}@amazon.com}

\begin{document}

\maketitle

\begin{abstract}
  Deploying computer vision models in Warehouse Facilities traditionally requires extensive resources for camera mounting, image collection, annotation, training, 
  and deployment - a process often needing repetition in each new environment due to camera mounting constraints and environmental variability. This paper explores an 
  innovative approach to streamline this process by conducting the standard procedure solely in a laboratory setting, focusing on vertical material handling systems
  and anomaly detection in forks of the systems. Through extensive experimentation, we have found that combining optimal camera placement, strategic image triggering, 
  careful model selection and model ensemble enables effective generalization from laboratory conditions to diverse warehouse facilities environments, potentially transforming warehouse automation 
  implementation by simplifying warehouse facilities deployment to just camera mounting, image collection, and model deployment, thereby saving significant resources and time typically spent on 
  image annotation and model retraining. \textbf{This is an experimental research study and not a production deployment.}
\end{abstract}

\begin{keywords}
Computer vision, warehouse automation, anomaly detection, generalization, material handling systems.
\end{keywords}

\section{Introduction}

Modern e-commerce companies operate hundreds of warehouse facilities globally, spanning multiple generations of technological advancement. The standard process of implementing computer vision models in these warehouses involves
five key steps: 1) mounting cameras, 2) collecting images, 3) annotating data, 4) training models, and 5) deployment. Despite
many computer vision models detecting identical features across multiple warehouse facilities, this process often requires repetition at each
facility due to two primary challenges: camera mounting constraints and environmental variability.
This paper examines computer vision model generalization from laboratory conditions to warehouse facilities environments, focusing on vertical material handling systems used for inter-floor transfer in modern automated warehouses. Component-level anomalies in such systems can impact operational efficiency. Our research specifically addresses anomaly detection in mechanical interface components (Fig 1) that facilitate material transfer. Structural deviations in these components can affect system performance (Fig 2).
Our study encompasses three facilities: laboratory, Facility A, and Facility B (operational warehouse facilities). While lab
provides controlled conditions for camera mounting, image capture, data annotation, and model training, Facility A and Facility B
present distinctly different environmental challenges affecting camera placement and model transferability. The key to
successful model generalization from laboratory to other warehouse facilities lies in maximizing image similarity across facilities. Our research
demonstrates that combining optimal camera positioning, strategic image triggering, appropriate model selection and model
ensemble techniques enables effective generalization from laboratory conditions to diverse warehouse facilities environments.


\begin{figure}
    \centering
    \begin{minipage}{0.5\textwidth}
        \centering
        \includegraphics[width=0.7\textwidth]{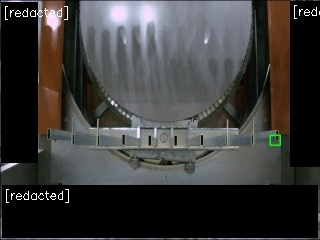} 
        \caption{Fork with area of interest in green square}
    \end{minipage}\hfill
    \begin{minipage}{0.5\textwidth}
        \centering
        \includegraphics[width=0.7\textwidth]{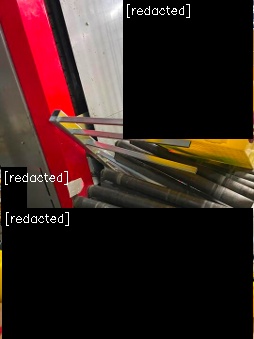} 
        \caption{Fork shuttle/roller crash}
    \end{minipage}
\end{figure}

\section{Related Work}

The challenge of deploying machine learning models across diverse operational environments has been extensively studied across multiple domains. Our work builds upon and extends several key research areas in model generalization and domain adaptation.

\subsection{Distribution Shift and Domain Adaptation}

Machine learning models frequently encounter performance degradation when deployed in environments that differ from their training conditions---a phenomenon known as distribution shift or covariate shift, where $P(X_{\mathrm{train}}) \neq P(X_{\mathrm{test}})$. [1] demonstrated this challenge in pharmaceutical research, showing how chemical compound series evolve over time, creating natural distribution changes that degrade model performance on test sets. This fundamental challenge extends across domains, with extensive research exploring its broader impacts [2], [3], and [4].

Domain adaptation techniques have emerged as a primary approach to address distribution shift. Traditional methods focus on learning domain-invariant representations or adapting models post-training to new target domains. However, these approaches typically require access to target domain data during training or fine-tuning phases. In contrast, our work explores a complementary strategy: engineering the data collection pipeline to minimize distribution shift at the source, thereby reducing the need for complex adaptation algorithms.

\subsection{Model Robustness and Generalization}

Beyond distribution shift, model generalization can be compromised by overfitting [5], [6], and [7], where models learn training-specific patterns that fail to transfer to new environments. Regularization techniques  and architectural choices  have been proposed to mitigate overfitting. More recently, research has examined how different model architectures exhibit varying degrees of robustness to distribution shift. [16] demonstrated that Transformer-based architectures exhibit superior robustness compared to convolutional neural networks (CNNs) when facing adversarial perturbations and out-of-distribution samples, a finding that directly informs our model selection strategy.

Domain generalization research extends these concepts by training models on multiple source domains to improve performance on unseen target domains. [8] analyzed this challenge in healthcare, showing how Electronic Health Record (EHR) models trained on several hospitals often fail when applied to new, unseen hospitals. This mirrors our challenge of deploying computer vision models across diverse warehouse facilities with varying environmental conditions.

\subsection{Adversarial Robustness}

Adversarial machine learning research,  has revealed that even small, carefully crafted perturbations can fool deep neural networks. While our work does not directly address adversarial attacks, the robustness principles developed in this domain—particularly the importance of model architecture selection and ensemble methods—inform our approach to handling natural environmental variations across facilities.

\subsection{Industrial Computer Vision Deployment}

Despite extensive theoretical research on domain adaptation and model generalization, practical frameworks for deploying computer vision systems across multiple industrial sites remain limited. Most existing work focuses on algorithmic improvements for post-deployment adaptation, requiring significant computational resources and labeled data at each new site. Our work addresses a critical gap by demonstrating that strategic engineering of the data collection process—through optimal camera placement, dynamic image triggering, and careful model selection—can enable effective cross-site deployment without site-specific retraining.

This approach represents a paradigm shift from adaptation-focused methods to **prevention-focused methods**, where generalization is achieved by maximizing visual consistency between training and deployment environments rather than by developing increasingly complex adaptation algorithms. The following sections detail our methodology for achieving this goal in the context of warehouse automation systems.

\section{Tools and techniques for computer vision model generalization from laboratory to warehouse facilities}
\label{headings}

To enhance model generalization from laboratory environments to warehouse facilities, we focus on minimizing the
discrepancy between training and test data distributions through various tools and techniques.

\subsection{Camera view selection}

The foundational element in ensuring model generalization is establishing consistency between training data (collected in
laboratory settings) and test data (from operational warehouse facilities). Camera view selection plays a crucial role in this alignment, as it
directly influences the visual characteristics of the collected data.
For fork anomaly detection, we evaluated several camera perspectives to determine the optimal viewing angle. Four distinct
camera views were tested in the laboratory environment, as illustrated in Figures 1, 3, and 4.
Evaluation of Camera Views:

Views 1-2 (Figures 3-4): These views presented significant challenges:
\begin{itemize}

\item Multiple forks appeared in a single frame, complicating the model's focus.

\item Excessive unrelated background elements increased the difficulty of achieving good model performance.
\end{itemize}

View 3 (Figure 1): This view, with the camera mounted on a floor-standing tripod, offered multiple
advantages:

\begin{itemize}

\item Single-fork focus: Only one fork appears in each frame, eliminating the need for the model to differentiate
between multiple forks.

\item Comprehensive tine visibility: The frontal view allows the model to assess both horizontal and vertical tine
positioning, enabling detection of abnormal tine spacing in both dimensions.

\item Background consistency: The frontal view increases the likelihood of capturing similar backgrounds (vertical material handling systems
wall) in both laboratory and warehouse facilities environments, enhancing generalization potential.
\end{itemize}

Based on these evaluations, View 3 was selected as the optimal camera perspective for fork anomaly detection, offering the
best balance of focused subject matter and potential for generalization across different environments.

\begin{figure}
    \centering
    \begin{minipage}{0.5\textwidth}
        \centering
        \includegraphics[width=0.9\textwidth]{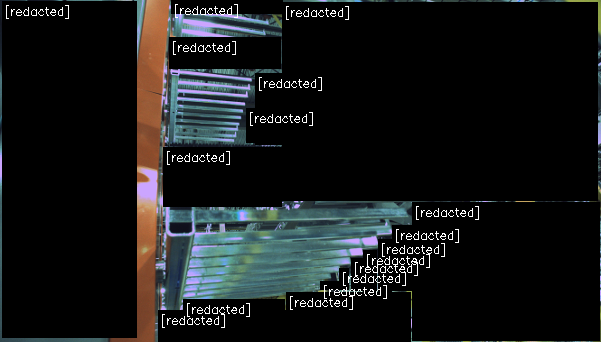} 
        \caption{Camera view 1}
    \end{minipage}\hfill
    \begin{minipage}{0.5\textwidth}
        \centering
        \includegraphics[width=0.9\textwidth]{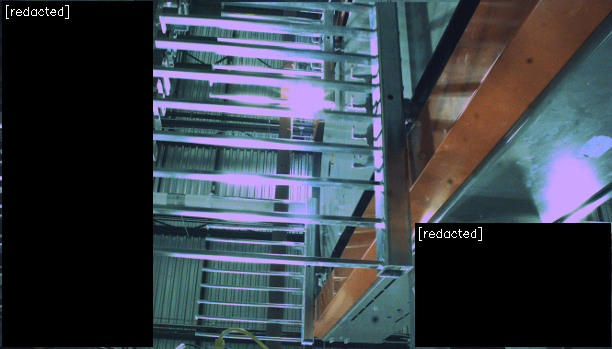} 
        \caption{Camera view 2}
    \end{minipage}
\end{figure}

\subsection{Image trigger}

Our approach to image capture involves a two-stage strategy that balances comprehensive model training with focused
inference:
(1) Training Phase: During model training, we expose the algorithm to images of the fork from various angles. This approach
ensures that the model develops a robust understanding of the fork's characteristics across different perspectives.
(2) Inference Phase: For the inference stage, we implement a more controlled image capture process to enhance similarity
between laboratory and warehouse facilities environments. 

This process involves:
(1) Area of Interest (AOI) Definition: We define a specific region, represented by a green square in our setup, as AOI. This targeted approach helps standardize the fork's position across different environments.
(2) Dynamic Image Capture: Our image trigger mechanism is based on frame differences within the AOI over a couple of seconds window.

The process works as follows:
(1) Continuous Monitoring: The system continuously analyzes frames within the AOI. (2) Movement Detection: When a fork
passes through the AOI, it creates significant frame-to-frame differences. (3) Trigger Activation: Upon detecting these
differences, the system triggers an image capture.

This dynamic triggering method ensures that: (1) Images are captured at consistent and relevant moments across different environments.
(2) The fork is positioned similarly in images from both laboratory and warehouse facilities settings.
(3) Irrelevant or non-informative frames are automatically filtered out.

By implementing this targeted image capture strategy, we significantly increase the likelihood of obtaining visually similar data
from both laboratory and warehouse facilities environments, thereby enhancing the model's generalization capabilities.

\subsection{Model selection and implementation}
Our approach to fork anomaly detection involves careful model selection and a two-stage detection process:
(1) Object Classification: We established two distinct classes for fork tine annotation (Fig 5): Front Rectangle and Front Stick.
Model Evaluation: We trained and evaluated multiple state-of-the-art computer vision models in the laboratory
environment: Mask R-CNN, RTMDet, Grounding Dino.
Each model was trained on a dataset of 5,866 images and evaluated on a test set of 1,467 images to achieve optimal mean Average Precision (mAP) before warehouse facilities deployment (Table 1). 
(2) Post-Processing Analysis: Following successful detection of Front Rectangle and Front Stick components, we
implemented a post-processing logic that: i) Analyzes horizontal spacing between tines. ii) Evaluates vertical alignment.
iii) Identifies anomalous configurations based on predefined spacing thresholds. 

This comprehensive approach combines robust object detection with geometric analysis to effectively identify fork anomalies
across different operational environments. Given the similar mAP across all classes for these three models, we anticipate comparable performance in fork anomaly detection.

\begin{table}
  \caption{Test set bounding box mAP for Front Rectangle and Front Stick detection in laboratory environment}
  \label{sample-table}
  \centering
  \begin{tabular}{llll}
    \toprule
    \cmidrule(r){1-2}
    Model     & All classes     & Front Rectangle & Front Stick \\
    \midrule
    Mask R-CNN & 94.90 \%  & 93.30 \% & 96.60 \%     \\
    RTMDet     & 95.30 \% & 93.20 \%  & 97.40 \%    \\
    Grounding Dino  & 96.10 \%  & 94.30 \%  & 97.80 \%  \\
    \bottomrule
  \end{tabular}
\end{table}

\begin{figure}
    \centering
    \begin{minipage}{0.5\textwidth}
        \centering
        \includegraphics[width=0.9\textwidth]{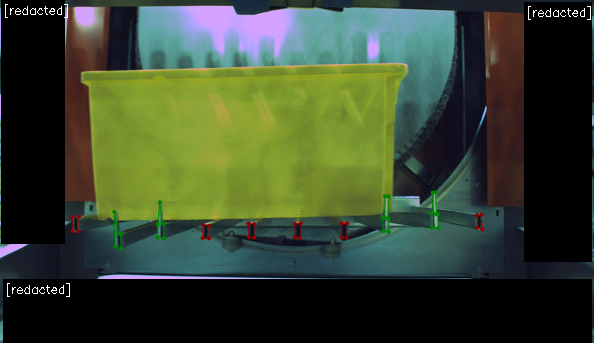} 
        \caption{Fork annotation; Red is front rectangle and green is front stick}
    \end{minipage}\hfill
    \begin{minipage}{0.5\textwidth}
        \centering
        \includegraphics[width=0.9\textwidth]{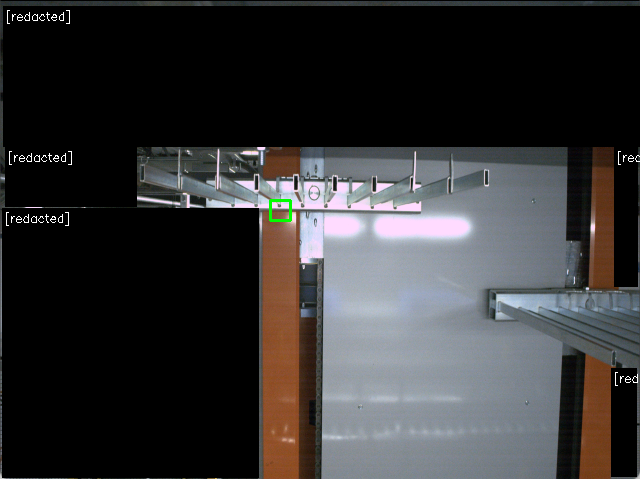} 
        \caption{Image trigger at Facility B}
    \end{minipage}
\end{figure}

\subsection{Model ensemble}
Model ensemble techniques are powerful tools for enhancing prediction accuracy and robustness. Common ensemble methos include 
Bagging (Bootstrap Aggregating) [13], Boosting [14] and Stacking (Stacked Generalization) [15].
These methods leverage multiple models to generate final predictions, often yielding superior performance compared to
individual models.

In our approach, we implemented a simple yet effective ensemble strategy using at least two models. This method offers
several advantages:

(1) Enhanced reliability: When one model detects an anomaly, a second model is used for verification. Consensus between
models increases confidence in the anomaly detection.

(2) Error mitigation: Individual models may produce false positives or false negatives due to incorrect detection of Front
Rectangles and Front Sticks. These errors can lead to inaccurate anomaly detection in the post-processing logic, which relies
on correct spatial relationships between components.

(3) Performance boost: By requiring agreement between at least two models, we significantly reduce the likelihood of false
anomaly detections. This approach effectively mitigates individual model weaknesses and enhances overall system reliability.

(4) Cost-effective improvement: Ensemble methods offer a relatively simple and economical way to improve model performance
without the need for extensive additional data collection or model retraining.
Our implementation demonstrates that even a basic ensemble approach can yield substantial improvements in anomaly
detection accuracy, particularly in challenging and variable environments like warehouse facilities.

\section{Experiments}
\label{others}

\subsection{Camera view implementation in warehouse facilities}

We conducted experiments to evaluate the performance of our computer vision model, trained in the laboratory, when
deployed at the Facility B. Upon arrival at Facility B, we encountered significant environmental differences that
necessitated adaptations to our camera setup:
\begin{itemize}

\item Environmental differences: 1. Facility B features two adjacent lifts, contrasting with our single-lift laboratory setup. 2. The left lift closely resembled our laboratory configuration, allowing for a similar floor-mounted, front-facing camera setup. 3. The right lift presented challenges: 
its lowest fork position was approximately 1 meter above floor level, rendering our standard floor-mounted setup ineffective.

\item Operational considerations: We learned that during maintenance operations may need to remove shuttles above the camera. This scenario could potentially impact floor-mounted
cameras, necessitating an alternative mounting solution.

\item Adapted camera mounting: To address these challenges, we devised a new camera mounting strategy: 1. We identified a mounting location similar to existing security camera positions. 2. This placement aimed to replicate, as closely as possible, the visual perspective achieved in our laboratory
setting.

\item New challenges: The adapted camera position introduced a new complexity: two forks now appear in a single frame.
This necessitated the development of additional post-processing logic to handle multiple fork detection and analysis.
\end{itemize}

These experiments highlight the importance of flexibility in camera placement strategies when transitioning from controlled
laboratory environments to diverse operational settings in warehouse facilities. Our ability to adapt our approach while
maintaining visual similarity to training data was crucial for successful model deployment

\subsection{Image trigger adaptation in warehouse facilities}

To maintain consistency with our laboratory training data while accommodating facility-specific configurations, we modified our
image trigger implementation:

\begin{itemize}
  \item Area of Interest (AOI) Adjustment: 1. We repositioned the AOI as shown in Fig 6. 2. This new location was strategically selected to capture forks at positions visually similar to our training data.
  \item Trigger Mechanism: 1. The system monitors the designated AOI for fork movement. 2. When a fork passes through this zone, the trigger activates and captures an image. 3. This timing ensures optimal fork positioning that closely matches our training dataset characteristics.
\end{itemize}

This adaptation demonstrates how careful positioning of the trigger zone can help maintain visual consistency between
laboratory and warehouse facilities environments, which is crucial for reliable model performance.

\subsection{Model selection and performance in warehouse facilities}
We evaluated three models trained in our laboratory environment—Mask R-CNN, RTMDet and Grounding DINO—on 1,089 images collected from
Facility B. The comparative performance is illustrated in Table 2. The key distinction between Grounding DINO and the other two models lies in their backbone architectures: Grounding DINO employs a Swin Transformer, whereas Mask R-CNN and RTMDet utilize convolutional-based architectures. Recent research has demonstrated that Transformer-based models exhibit greater robustness compared to CNNs [16].

\begin{table}
  \caption{Fork anomaly detection at Facility B}
  \label{sample-table}
  \centering
  \begin{tabular}{lll}
    \toprule
    \cmidrule(r){1-2}
    Model     & False Postive Rate (FPR)     & Accuracy \\
    \midrule
    Mask R-CNN & 36.64 \%  & 63.36 \%     \\
    RTMDet     & 96.79 \% & 3.21 \%      \\
    Grounding Dino     & 6.61 \%       & 93.39 \%  \\
    \bottomrule
  \end{tabular}
\end{table}

\subsection{Model ensemble implementation in warehouse facilities}

Analysis of Prediction Patterns: Our testing generated 1,089 total predictions from 20 forks in a single lift, averaging approximately 54 predictions per fork—comprising 4 anomaly predictions and 50 normal state predictions.

Current Limitations: (1) Lack of unique fork identification capability (2) Inability to track predictions across specific forks over time. 

Proposed Enhancement through model ensemble: We propose implementing a temporal ensemble approach once fork identification is integrated:

(1) Fork-Level Aggregation: i) Track predictions for each unique fork. ii) Aggregate multiple predictions over time. 
iii) Require minimum threshold of anomaly detections. 

(2) Decision Logic: i) Implement a "two-strike" rule ii) Classify fork as anomalous only after two or more anomaly predictions
iii) Consider temporal distribution of anomaly predictions

(3) Expected Benefits: i) Reduced False Positive Rate (FPR). After a normal fork is predicted anomaly, the expected probability of its predicted anomaly again is 4/54 = 7.4\%. 
  So 2nd prediction has 92.6\% chance to correct the 1st wrong prediction. In theory, the FPR will be reduced from 6.61\% to 0\%
  ii) Increased overall accuracy. Follow the same logic as the reduced false positive rate, the overall accuracy will increase from 93.49\% to 100\%

\section{Limitations}

Our implementation faced several significant challenges, particularly at the Facility A:

(1) Physical Mounting Constraints: i) Existing wire infrastructure (Fig 7) prevents floor-level camera mounting
ii) Limited options for alternative mounting positions near floor level

(2) Camera Technology Limitations: We attempted to overcome mounting constraints using fisheye cameras mounted on
the wire, but encountered several issues: i) Peripheral Front Rectangles (far end, both sides) remained obscured even after image undistortion (Fig 9,
10). ii) Critical components of the fork structure were not consistently visible

(3) Background Complexity: i) Far-end Front Rectangles appeared against highly variable backgrounds
ii) Estimated requirement of 100,000+ training images to adequately capture background variation. While a lot less data requirement method exists, alternative method is preferred

(4) Alternative Top-of-lift Mounting Challenges: i) Environmental obstacles near lift structure
ii) Difficulty achieving proper frontal view of forks (Fig 8). 
These limitations highlight the challenges of transitioning from controlled laboratory conditions to diverse operational
environments, particularly when dealing with existing infrastructure constraints and varying environmental conditions.

\begin{figure}
    \centering
    \begin{minipage}{0.5\textwidth}
        \centering
        \includegraphics[width=0.9\textwidth]{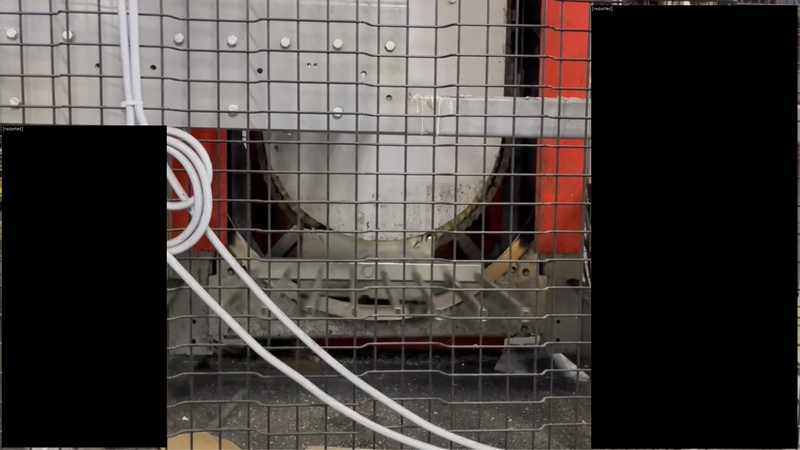} 
        \caption{Facility A lift looking from on the floor}
    \end{minipage}\hfill
    \begin{minipage}{0.5\textwidth}
        \centering
        \includegraphics[width=0.9\textwidth]{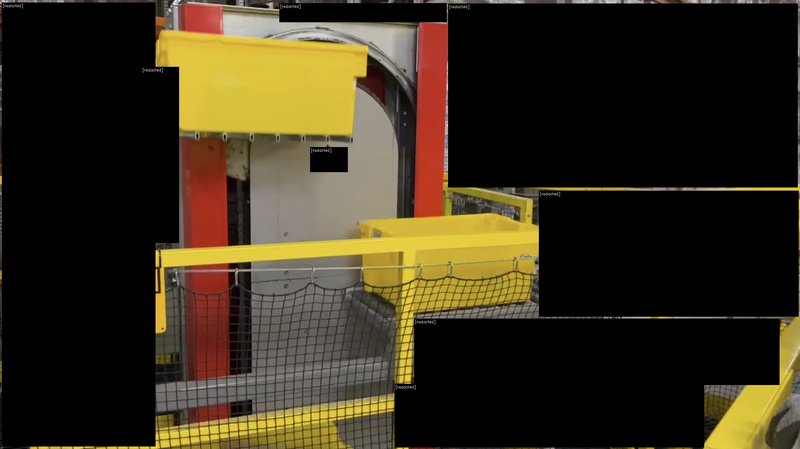} 
        \caption{Facility A lift looking from the top}
    \end{minipage}
\end{figure}

\begin{figure}
    \centering
    \begin{minipage}{0.5\textwidth}
        \centering
        \includegraphics[width=0.9\textwidth]{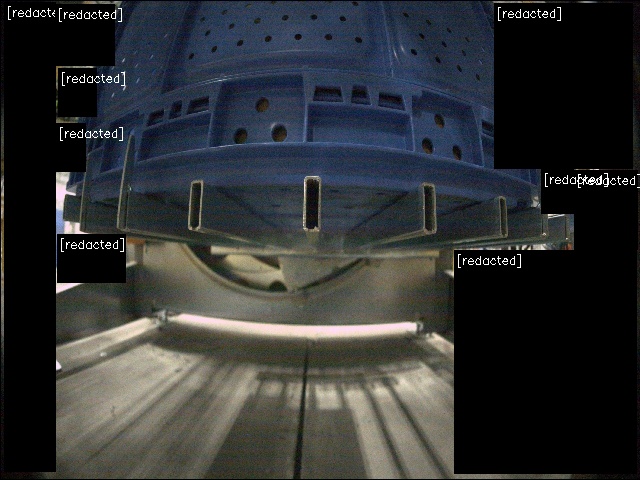} 
        \caption{Image captured with fish eye camera in laboratory}
    \end{minipage}\hfill
    \begin{minipage}{0.5\textwidth}
        \centering
        \includegraphics[width=0.9\textwidth]{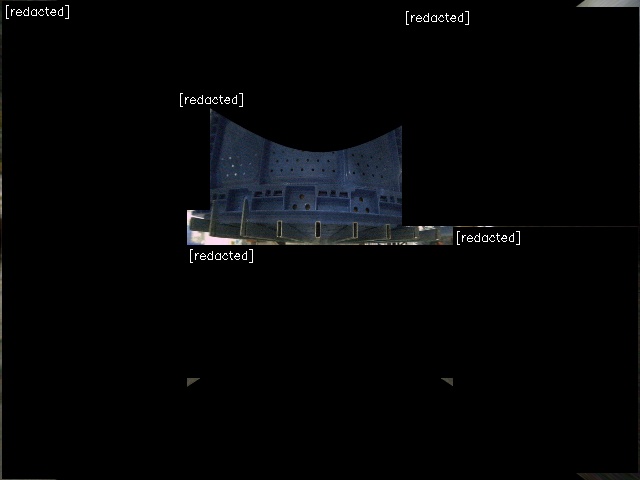} 
        \caption{Undistorted fish eye image in laboratory}
    \end{minipage}
\end{figure}

\section{Conclusions}

This study demonstrates the potential for deploying computer vision models for anomaly detection in warehouse facilities, focusing on a vertical tote lift system. Our multi-faceted approach enhances model generalization from laboratory settings to diverse facility environments.

Key findings include:

\begin{itemize}

\item Optimal Camera Placement: A frontal fork view, achieved through strategic camera positioning, significantly improves model generalizability.

\item Dynamic Image Triggering: Our Area of Interest (AOI) based triggering mechanism ensures consistent image capture across environments, enhancing data similarity between training and deployment settings.

\item Model Selection: Comparative analysis demonstrated that careful model selection (e.g., Grounding DINO) dramatically improves out-of-the-box performance in new environments.

\item Ensemble Techniques: A temporal ensemble approach reduces false positives and increases accuracy, particularly when combined with unique fork identification.

\item Adaptability: Experiments across different facilities underscored the importance of flexible implementation strategies to overcome site-specific challenges.

\end{itemize}

Despite these advancements, limitations arose in environments with complex infrastructure constraints, highlighting areas for future work:

\begin{itemize}
  \item Development of robust camera mounting solutions for varied warehouse layouts
  \item Advanced image processing techniques to overcome visibility issues in constrained spaces
  \item Transfer learning and domain adaptation methods to further enhance model generalization
\end{itemize}

In conclusion, our research presents a promising framework for streamlining computer vision deployment in warehouse automation. By reducing the need for extensive on-site data collection and retraining, this approach enhances the efficiency and scalability of anomaly detection across warehouse facilities, with potential for broader industrial applications.

\section*{References}
\medskip

\small

[1] G. McGaughey, W.P. Walters, and B. Goldman, "Understanding covariate shift in model performance," F1000Research, vol.5(Chem Inf Sci):597, 2016. doi: 10.12688/f1000research.8317.3

[2] "Practical Issues in Data Science Part 2: Distribution Shift (Part 1)," Medium. [Online]. Available: \url{https://medium.com/analytics-vidhya/practical-issues-in-data-science-part-2- \
distribution-shift-part-1-416754c01905}

[3] A. Singh, "Detecting and Mitigating Data Distribution Shift," GitHub. [Online]. Available: \url{https://singhay.github.io/machine\%20learning/data-shift/}

[4] "Domain adaptation," Wikipedia. [Online]. Available: \url{https://en.wikipedia.org/wiki/Domain_adaptation}

[5] C. Zhang, S. Bengio, M. Hardt, B. Recht, and O. Vinyals, "Understanding deep learning requires rethinking generalization," arXiv preprint arXiv:1611.03530, 2016.

[6] H. Zou and T. Hastie, "Regularization and Variable Selection via the Elastic Net," Department of Statistics, Stanford University, Technical Report, Dec. 2003 (revised Aug. 2004).

[7] I. Goodfellow, Y. Bengio, and A. Courville, Deep Learning. MIT Press, 2016.

[8] A. Abdel Hai, M.G. Weiner, A. Livshits, J.R. Brown, A. Paranjape, W. Hwang, L.H. Kirchner, N. Mathioudakis, E.K. French, Z. Obradovic, and D.J. Rubin,
"Domain generalization for enhanced predictions of hospital readmission on unseen domains
among patients with diabetes," Artificial Intelligence in Medicine, vol. 158, p. 103010, Dec. 2024.

[9] I. J. Goodfellow, J. Shlens, and C. Szegedy, "Explaining and harnessing adversarial examples," arXiv preprint arXiv:1412.6572, 2014.

[10] "Adversarial machine learning," Wikipedia. [Online]. Available: \url{https://en.wikipedia.org/wiki/Adversarial_machine_learning}

[11] X. Yuan, P. He, Q. Zhu, and X. Li, "Adversarial examples: Attacks and defenses for deep learning," IEEE Transactions on Neural Networks and Learning Systems, vol. 30, no. 9, pp. 2805-2824, 2019.

[12] A. C. Serban, E. Poll, and J. Visser, "Adversarial examples - A complete characterisation of the phenomenon," arXiv preprint arXiv:1810.01185, 2018.

[13] L. Breiman, "Bagging predictors," Machine Learning, vol. 24, no. 2, pp. 123-140, 1996.

[14] R. E. Schapire, "The strength of weak learnability," Machine Learning, vol. 5, no. 2, pp. 197-227, 1990.

[15] D. H. Wolpert, "Stacked generalization," Neural Networks, vol. 5, no. 2, pp. 241-259, 1992.

[16]  Y. Bai, J. Mei, A. L. Yuille, and C. Xie, "Are Transformers more robust than CNNs?" in {\it Proc. Advances in Neural Information Processing Systems (NeurIPS)}, vol. 34, pp. 26831--26843, 2021.

\end{document}